\documentclass{article}

    \PassOptionsToPackage{numbers, compress}{natbib}
\usepackage[preprint]{neurips_2026}
\usepackage[utf8]{inputenc}
\usepackage[T1]{fontenc}
\usepackage{hyperref}
\usepackage{url}
\usepackage{booktabs}
\usepackage{amsfonts}
\usepackage{amsmath}
\usepackage{nicefrac}
\usepackage{microtype}
\usepackage[table]{xcolor}
\usepackage{graphicx}
\usepackage{multirow}
\usepackage{enumitem}
\usepackage{bbm}
\usepackage{makecell}
\usepackage{xspace}

\newcommand{\method}{\textbf{ReGenHuman}\xspace}

\title{\method: Re-Generating Human Appearances for Realistic Full-Body Video Anonymization}

\author{Adam Sun\thanks{Corresponding author.}, \space Eshaan Barkataki, Arnold Milstein, Gordon Wetzstein, Ehsan Adeli$^*$ \\
  Stanford University, Stanford, CA, USA \\
  \texttt{\{adsun,eadeli\}@stanford.edu} \\
  \url{https://regenhuman.github.io}
}

\begin{document}

\maketitle

%%%%%%%%%%%%%%%%%%%%%%%%%%%%%%%%%%%%%%%%%%%%%%%%%%%%%%%%%%%%%%%%%%%%%%%%%%%%%%%%
\begin{abstract}
Anonymizing human-centric video data is an understudied problem. Prior anonymization techniques either blur or redact pixels at the cost of realism and downstream utility, or generate frame-by-frame at the cost of temporal coherence. We introduce \method, the first full-body video anonymization pipeline that is simultaneously realistic, temporally consistent, and anonymous by construction. Contrary to past approaches which redact or edit the inputs directly, we propose a \emph{regenerate, don't edit} paradigm. Our approach composites 2D pose, segmentation, and monocular depth into two complementary conditioning streams---\textbf{StructAll} and \textbf{StructHuman}, which are used to fine-tune a video-to-video diffusion backbone on in-the-wild human videos, synthesizing the human regions entirely from identity-free structural cues. We evaluate our model on privacy, quality, and utility, and show that our \method achieves the best tradeoff across all three axes against current baselines. We further show that our anonymized videos remain effective for downstream tasks, including video question answering.
\end{abstract}

\section{Introduction}
\label{sec:intro}

Modern vision systems rely on large, human-centric video datasets to power applications ranging from action recognition and video understanding~\cite{caba2015activitynet, diba2020large, kay2017kinetics} to autonomous vehicles \cite{gawande2020pedestrian, zhang2022learnable, zhang2025more, zhang2020investigating} and video surveillance~\cite{ye2021deep, zheng2017person, vijeikis2022efficient, sultani2018real}. Data is especially important in clinical settings, where bedside monitoring and ambient intelligence applications require diverse, multi-person footage captured in real-world environments~\cite{mopidevi2025medviddeid, zhu2020deepfakes, dai2025developing, cai2025safetriage, ahmedt2024deep, wang2025medgen}. However, raw videos present a severe privacy risk; identity cues are apparent not only in faces and bodies but also in clothing, gait, and background surroundings~\cite{zhang2025more, zwick2024context, weiss2025privacy}. Consequently, regulations such as the US Health Insurance Portability and Accountability Act (HIPAA)~\cite{annas2003hipaa} and the General Data Protection Regulation~\cite{voigt2017eu} require consent that is effectively impossible to obtain at the scale. This challenge is compounded by the fact that trained models themselves are vulnerable to privacy breaches. Attacks such as membership inference can recover identifying frames from a model's weights alone \cite{shokri2017membership, li2025vid, hu2022m}.

\begin{figure}[t]
  \includegraphics[width= \linewidth]{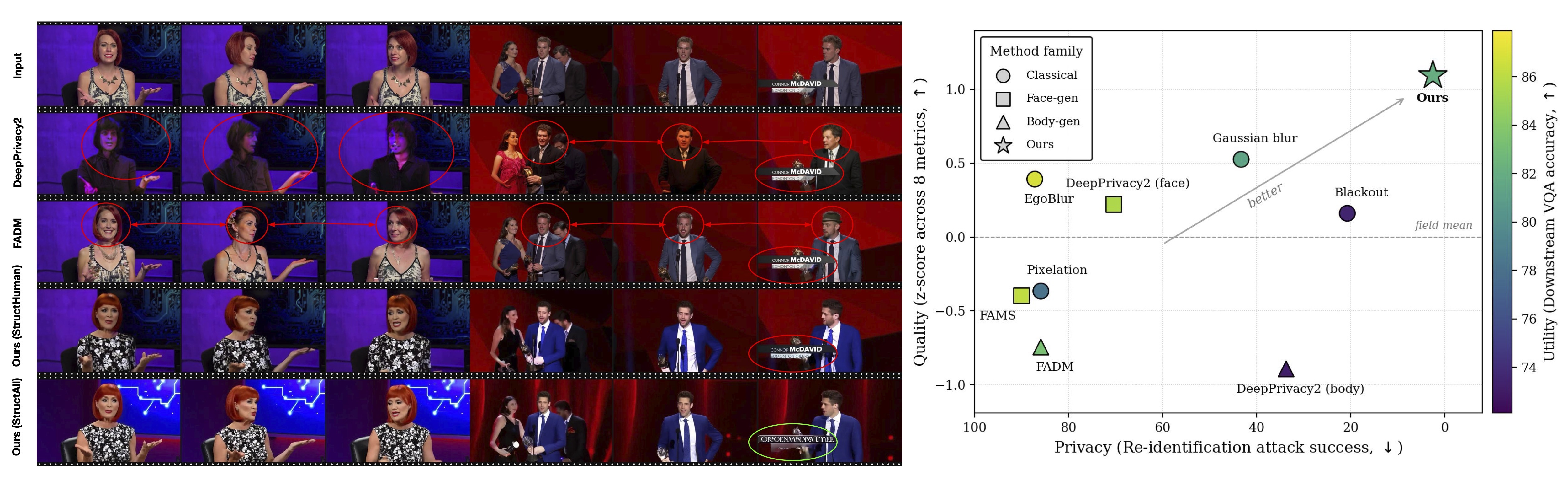}
  \caption{We propose \method, the first pipeline for realistic full-body video anonymization. Past methods for full-body anonymization such as DeepPrivacy2~\cite{hukkelaas2023deepprivacy2} and FADM~\cite{zwick2024context} are inadequate for video --- they either generate anatomically unrealistic humans or do not depict the same identity across frames. On the other hand, our \textbf{ReGenHuman}'s generations are realistic and temporally consistent, with our \textbf{StructAll} generation being able to anonymize the background and accompanying text (\emph{left}). We evaluate current anonymization approaches on privacy (Clothes-changing re-identification mAP on CCVID~\cite{gu2022clothes} with CSCI~\cite{pathak2025colors}), quality (z-score across metrics from VBench 1 and 2~\cite{huang2024vbench, zheng2025vbench}), and utility (zero-shot video question answering accuracy on HOI-Gen1M~\cite{liu2025hoigen}). Our approach achieves the best tradeoff of all prior baselines, setting a new standard in video anonymization (\emph{right}). }
  \label{fig:teaser}
\end{figure}

The sensitive domains that stand to benefit most from large-scale video learning—such as hospitals, classrooms, and workplaces—are precisely the areas where data cannot be openly released. As a result, AI performance in these critical settings falls behind \cite{he2024foundation, sharma2021edunet}. The natural solution is anonymization: transforming a video so that no individual can be re-identified, while preserving the semantic content that downstream models depend on. Text~\cite{hassan2019automatic, mamede2016automated, pilan2022text} and image~\cite{hukkelaas2019deepprivacy, hukkelaas2023deepprivacy2, hukkelaas2023does, dinh2026unsafe2safe} anonymization have been extensively studied; however, video anonymization remains underexplored.

Following past work \cite{dinh2026unsafe2safe,zhang2025more}, we argue that a good video anonymizer must satisfy three objectives \emph{simultaneously}:
%\begin{itemize}
%\item 
\textbf{(P1) Privacy.} The output must not be re-identifiable by any means, including through face, body, clothing, and background cues.
%\item 
\textbf{(P2) Quality.} The video must be temporally and geometrically consistent, and realistic enough that models trained on real-world videos can utilize it without distribution shift.
%\item 
\textbf{(P3) Utility.} Actions, poses, human--object interactions, and scene context must remain recognizable, preserving the semantic structure required by downstream tasks.
%\end{itemize}

The past paradigm for anonymization mainly involves \emph{redacting, editing or perturbing} the original pixels---a paradigm that either removes useful information or does not guarantee full anonymization coverage. We propose the opposite paradigm: \emph{regenerate} the human region from scratch, conditioned only on identity-free structural signals. 

Our ReGenHuman pipeline consists of three distinct steps. Firstly, we extract pixel-precise human masks ~\cite{carion2025sam}, 2D pose skeletons ~\cite{yang2023effective}, monocular depth~\cite{chen2025video}, and a detailed text caption from the input video~\cite{liu2025hoigen,bai2025qwen3}. Secondly, we combine the visual conditionings into two structural conditioning videos \textbf{StructAll} and \textbf{StructHuman}, which are fully anonymous and preserve the context and dynamics of the original video. Lastly, we fine-tune a pretrained video-to-video diffusion model~\cite{wan2025wan,jiang2025vace} on real-world human videos from the HOIGen-1M dataset \cite{liu2025hoigen}. At inference, our trained anonymizer produces realistic, context-consistent anonymizations of input videos that are still useful for downstream applications, outperforming past anonymization approaches in privacy, quality, and utility.

Our contributions are the following:
\begin{itemize}[leftmargin=*,itemsep=1pt,topsep=2pt]
  \item We present the first realistic and temporally consistent video anonymization pipeline. By jointly denoising full clips conditioned on structural overlays, our method produces anonymized video that is visually realistic and contextually consistent with the input video.
  \item We propose a new anonymization paradigm: regenerate, don't edit. Unlike all prior methods, which edit, redact, or perturb the original pixels, we regenerate the human region from identity-free structural inputs --- our \textbf{StructAll} and \textbf{StructHuman}. This paradigm shift carries an architectural privacy guarantee: no original human pixel can reach the output.
  \item We outline a unified three-axis evaluation protocol for full-body video anonymization, jointly measuring \textbf{Privacy} (identity and text similarity, clothes-changing person re-identification), \textbf{Quality} (VBench- and VBench-2.0--aligned consistency, realism, and smoothness metrics), and \textbf{Utility} (video question answering).
\end{itemize}

We conduct detailed experiments showing that our method achieves the best privacy--quality--utility tradeoff among current anonymization approaches, including in the clinical domain. We hope this work inspires further research in the field of video anonymization.

%%%%%%%%%%%%%%%%%%%%%%%%%%%%%%%%%%%%%%%%%%%%%%%%%%%%%%%%%%%%%%%%%%%%%%%%%%%%%%%%
\section{Related Work}
\label{sec:related}

\begin{table}[t]
\centering
\small
\setlength{\tabcolsep}{6pt}
\renewcommand{\arraystretch}{1.1}
\newcommand{\yes}{\checkmark}
\newcommand{\no}{$\times$}
\caption{Comparing current video anonymization methods.
\textbf{Fully Private}: full-body coverage \emph{and} an architectural guarantee that no original human pixel reaches the output.
\textbf{Temporal}: output is temporally consistent (no per-frame flicker of identity). \textbf{Utility}: action, pose, HOI, and scene context are preserved.
\textbf{Realistic}: output is a human-viewable video that is photorealistic and in-distribution for models trained on regular video.
Ours is the only method that satisfies all four.}
\label{tab:method_comparison}
\begin{tabular}{lcccc}
\toprule
Method & Fully Private & Temporal & Utility & Realistic \\
\midrule
EgoBlur~\cite{raina2023egoblur}                                                                        & \no  & \yes & \yes  & \no \\
DeepPrivacy~\cite{hukkelaas2019deepprivacy}, FAMS~\cite{kung2025face}                                   & \no  & \no  & \yes & \no \\
DeepPrivacy2~\cite{hukkelaas2023deepprivacy2} & \yes  & \no  & \yes & \no  \\
FADM~\cite{zwick2024context}, Unsafe2Safe~\cite{dinh2026unsafe2safe} & \no  & \no  & \yes & \yes  \\
STPrivacy~\cite{li2023stprivacy}, Pixels2Privacy~\cite{aslam2026pixels}                                  & \no & \yes & \yes  & \no  \\
\midrule
\textbf{ReGenHuman (ours)}                                                                              & \yes & \yes & \yes & \yes \\
\bottomrule
\end{tabular}
\end{table}

\subsection{Image anonymization}
\label{sec:related:image}
The classical approach for image anonymization detects faces or bodies and blurs, pixelates, or masks out the regions. These pipelines remain the de-facto standard in deployed systems~\cite{raina2023egoblur} and are simple and verifiable. However, once faces, skin, and clothing are redacted, action recognition, video question answering, and other downstream tasks degrade sharply~\cite{hukkelaas2023does,zhang2025more}.

To improve the utility of anonymized images, generative anonymization instead replaces the redacted region with a synthesized one. For faces, GAN-based methods~\cite{hukkelaas2019deepprivacy,hukkelaas2023deepprivacy2,rosberg2023fiva,klemp2023ldfa} inpaint a new face from a bounding box or mask, and diffusion-based approaches such as Face Anonymization Made Simple (FAMS)~\cite{kung2025face} generate a plausible by face by partially noising and denoising the face region. These approaches may produce photorealistic faces but are insufficient for privacy: clothing, tattoos, and background remain intact, and modern body re-identification~\cite{gu2022clothes,pathak2025colors,zhou2019omni} re-links subjects from these cues even when the face is perfectly hidden~\cite{zhang2025more}.

A smaller line of work extends generative anonymization to the full body. DeepPrivacy2~\cite{hukkelaas2023deepprivacy2} inpaints the human region from a mask, but its GAN-based generations frequently look unrealistic. FADM~\cite{zwick2024context} uses a text-to-image diffusion model, but conditions the diffusion process on the original image and so offers no anonymization guarantee. Unsafe2Safe~\cite{dinh2026unsafe2safe} proposes a multi-stage pipeline combining LLMs, VLMs, and diffusion image editors. However, the editor stage suffers from the same issue, receiving the original image and relying on an edit instruction.

Furthermore, the image anonymizers above do not transfer well to video. With no access to other frames, the synthesized identity drifts over time, evident in Figure~\ref{fig:teaser}(\emph{left}). 

\subsection{Video anonymization}
\label{sec:related:video}
Video-native methods are sparse in the literature. Tracked-redaction approaches such as FAKER~\cite{ban2024faker} and the medical-domain MedVidDeID~\cite{mopidevi2025medviddeid} blur or replace the person consistently across frames, avoiding flicker but inheriting the utility loss of pixel-level redaction. Privacy-preserving action recognition~\cite{li2023stprivacy,aslam2026pixels} sidesteps pixel-level anonymization entirely by removing identity-bearing tokens from the input, but the output is not a human-viewable video and cannot be released for general downstream use. To the best of our knowledge, no generative video anonymization method exists.

\subsection{Controllable video generation}
\label{sec:related:ctrlvid}
Text-to-video diffusion models~\cite{blattmann2023stable,hacohen2024ltx,yang2024cogvideox,wan2025wan} denoise an entire clip jointly, producing temporally coherent video from a text prompt. Subsequent work added structural conditioning---depth, 2D pose, segmentation masks, or Canny edges as per-frame control channels~\cite{yang2018pose,ma2024follow,xing2024make,xi2026omnivdiff,wan2025wan}---and unified video-to-video models such as Wan Fun Control~\cite{wan2025wan} and VACE~\cite{jiang2025vace} accept an arbitrary mix of structural controls, masks, plus an optional reference image.

No prior anonymization method can anonymize a video consistently and realistically across frames. Furthermore, controllable video diffusion models were developed for creative generation and, to our knowledge, have never been applied for anonymization. Yet their two core properties---\emph{joint temporal denoising} and \emph{appearance-free structural conditioning}---are exactly the features that every prior video anonymizer was missing. We combine these insights and propose utilizing video-to-video diffusion models as full-body anonymizers.
\section{Method}
\label{sec:method}

\begin{figure}[t]
  \centering
  \includegraphics[width=\linewidth]{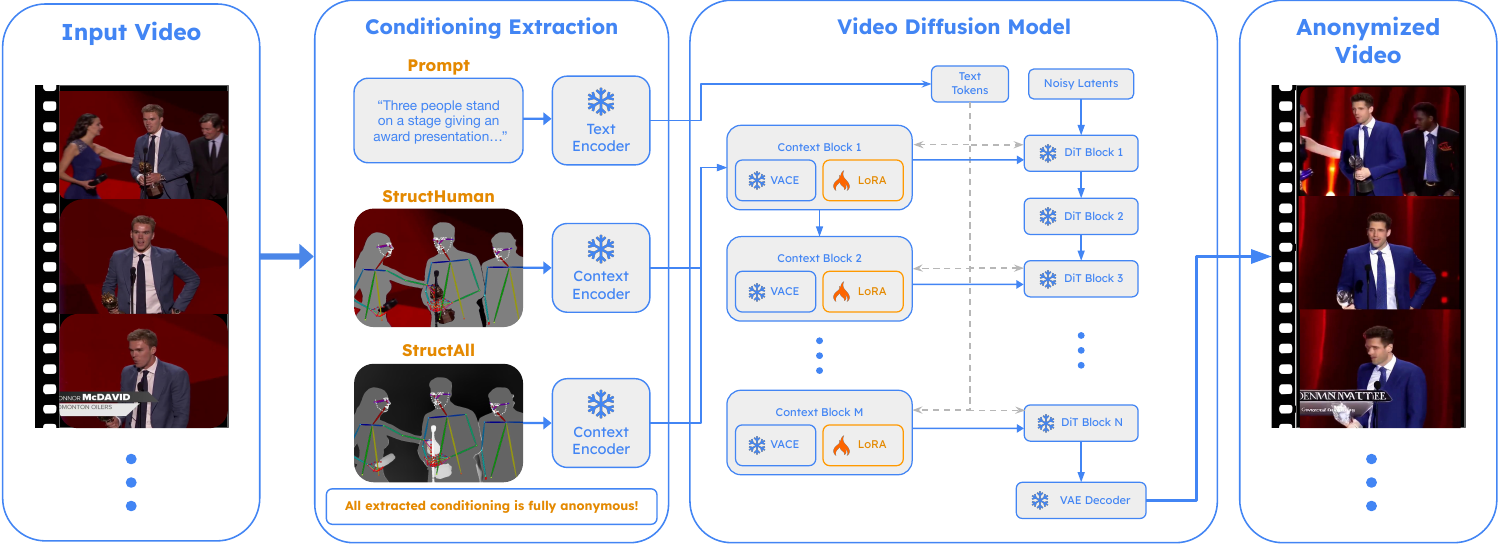}
  \caption{\textbf{ReGenHuman pipeline.} Given an input video, we extract completely anonymous conditioning --- a text caption in addition to our StructHuman or StructAll videos. We freeze both the DiT blocks and context blocks of VACE~\cite{jiang2025vace} and fine-tune LoRA~\cite{hu2022lora} inside each Context Block. At inference, we produce an anonymization \emph{without the model directly seeing the input video}.}
  \label{fig:pipeline}
\end{figure}

\subsection{Wan VACE}
\label{sec:bg}

Wan~\cite{wan2025wan} is an open-source video diffusion model that has been pretrained on a large corpus of video data, making it generalizable for our use case. Let $\mathbf{x}=(x^1,\dots,x^T)$ denote a video clip with 3D-VAE latent $\mathbf{z}_0=\mathcal{E}(\mathbf{x})$~\cite{wan2025wan}. VACE~\cite{jiang2025vace} extends Wan for unified controllable video generation. 

VACE works through two constructs. First, a \emph{Video Condition Unit} (VCU) packs task conditioning into a single video-shaped tensor $\mathbf{c}=(c^1,\dots,c^T)\in\mathbb{R}^{T\times H\times W\times 3}$ that may carry an arbitrary mix of structural signals (depth, pose, mask). Second, a stack of \emph{Context Blocks} blocks run in parallel to the frozen DiT backbone, encoding the VCU with the same 3D VAE and injecting hidden states into every DiT layer. The rectified-flow denoiser becomes $v_\theta(\mathbf{z}_t,t,\mathcal{E}(\mathbf{c}),y)$ with text prompt $y$. The key property for anonymization is that the VCU is injected per-pixel, so we can hand the model a heterogeneous overlay without modifying the architecture.

\subsection{Conditioning signal extraction}
\label{sec:conditioning}
\begin{figure}[t]
  \centering
  \includegraphics[width=\linewidth]{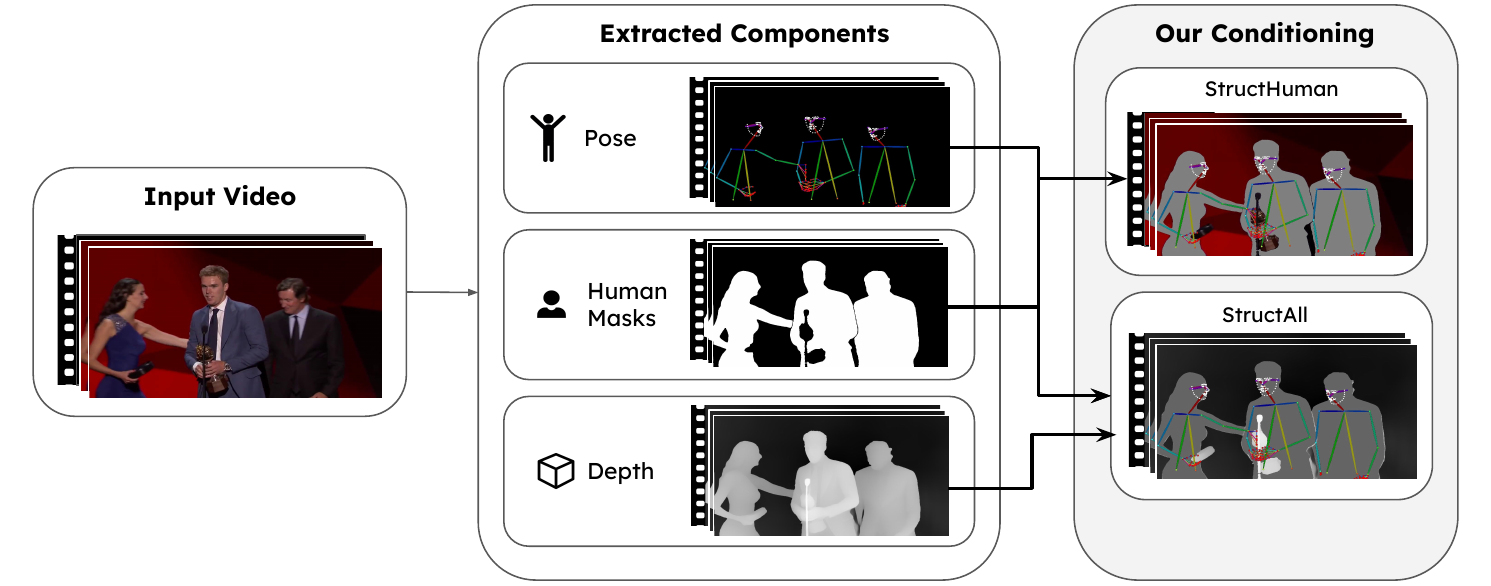}
  \caption{\textbf{Conditioning extraction.} We propose two types of conditioning – StructHuman and StructAll. Both conditioning videos anonymize the human with a pose overlaid on top. StructHuman reveals the original background and objects, while StructAll uses the extracted depth as background.}
  \label{fig:conditioning}
\end{figure}

From an input clip $\mathbf{x}$, we extract three per-frame signals as full-length videos plus a clip-level caption $y$. Using state-of-the-art off-the-shelf models, we obtain a human-instance mask video $\mathbf{M}\in\{0,1\}^{T\times H\times W}$ from SAM3~\cite{carion2025sam}, which propagates instance identities across frames; a rendered DWPose~\cite{yang2023effective} skeleton video $\mathbf{P}$; and a monocular depth video $\mathbf{D}$ from Video Depth Anything~\cite{chen2025video}, whose temporal-attention layers enforce frame-to-frame depth consistency.

Each of these three signals plays a distinct role and is appearance-free by construction. The human mask $\mathbf{M}$ tells the generator \emph{where} a person must appear, outlining the exact region to generate. The depth map $\mathbf{D}$ anchors the person in the 3D scene and provides the necessary geometry and context of the scene without revealing identifying information such as text. Pose $\mathbf{P}$ ensures that the regenerated person is positioned in the same way as the source. Furthermore, the descriptive caption $y$ ensures that the overall context of the clip remains consistent with the original.

We compose our signals into two conditioning videos. With $\Pi^t$ denoting the pixels occupied by the poses, we define:

% TODO(figure): Conditioning formulation figure.
% A single panel (or 2x3 grid) that visualizes, for one representative frame x^t:
%   (a) original frame x^t
%   (b) human mask M^t              (c) pose P^t + stroke support \Pi^t
%   (d) depth D^t                    (e) neutral-gray fill g / mean-depth fill \mu(D^t)
%   (f) resulting StructAll c^t_All  (g) resulting StructHuman c^t_Hum
% with arrows annotating Eq.~\ref{eq:structall} and Eq.~\ref{eq:structhuman}.
% Purpose: show readers at a glance that inside M^t there are only pose strokes / flat fills,
% and that StructAll vs StructHuman differs only in what fills the non-human background.

%\begin{itemize}
%\item
{\textbf{StructHuman}, which keeps non-human pixels (such as held objects and the background) non-anonymized when they carry no private information (e.g., public streets, generic hospital rooms). This approach trades scene anonymization for richer context.
\begin{equation}
  c^t_{\mathrm{Hum}} \;=\; \Pi^t\odot P^t \;+\; (1-\Pi^t)\odot\Big[\,M^t\odot \mathbf{g} \;+\; (1-M^t)\odot x^t\,\Big].
  \label{eq:structhuman}
\end{equation}}

%\item 
{\textbf{StructAll}, which uses depth as the background, yielding a fully-anonymized frame (no original pixels anywhere) while preserving scene geometry and the subject's pose:
\begin{equation}
  c^t_{\mathrm{All}} \;=\; \Pi^t\odot P^t \;+\; (1-\Pi^t)\odot\Big[\,M^t\odot \mathbf{g} \;+\; (1-M^t)\odot D^t\,\Big].
  \label{eq:structall}
\end{equation}}

%\end{itemize}
% \begin{figure}[t]
%   \centering
%   \fbox{\rule[-0.5cm]{0cm}{4cm}\rule[-0.5cm]{12cm}{0cm}}
%   \caption{\textbf{TODO: Conditioning formulation figure.} Visualizes the construction of $c^t_{\mathrm{All}}$ (Eq.~\ref{eq:structall}) and $c^t_{\mathrm{Hum}}$ (Eq.~\ref{eq:structhuman}) from $x^t$, $M^t$, $P^t$, and $D^t$, making the human-region invariant (Eq.~\ref{eq:invariant}) visible at a glance.}
%   \label{fig:conditioning}
% \end{figure}
In both cases the human regions are grayed out with $\mathbf{g}= (128,128,128)$ before the pose is overlayed on top. This prevents identifiable human features from leaking through the depth map. Notice:
\begin{equation}
  \forall t,\;\;\big[c^t\odot M^t\big]\;\text{depends on}\;P^t\;\text{only, not on}\;x^t,
  \label{eq:invariant}
\end{equation}
holds by construction, providing an architectural privacy guarantee: the diffusion model has no path through which the original human appearance can flow to the output. 

We keep the entire video-to-video diffusion model---both the Wan DiT backbone and the pretrained VACE Context Block weights---\emph{frozen}, and add rank-$\rho{=}32$ LoRA~\cite{hu2022lora} on the projection matrices of every Context Adapter block. This allows us to utilize the high-quality conditional generation capabilities of the pretrained model without dealing with catastrophic forgetting \cite{kirkpatrick2017overcoming}. We fine-tune with the standard conditional rectified-flow loss on $(\mathbf{x},\mathbf{c},y)$ triples, with $\mathbf{c}\in\{\mathbf{c}_{\mathrm{All}},\mathbf{c}_{\mathrm{Hum}}\}$ depending on the adapter. More implementation details can be found in the appendix.

%%%%%%%%%%%%%%%%%%%%%%%%%%%%%%%%%%%%%%%%%%%%%%%%%%%%%%%%%%%%%%%%%%%%%%%%%%%%%%%%x
\section{Experiments}
\label{sec:experiments}

\subsection{Setup}
\label{sec:setup}
For training our model, we utilize a randomly sampled 7{,}500-clip subset of HOIGen-1M~\cite{liu2025hoigen}, a corpus of over one million in-the-wild human--object interaction (HOI) videos curated from diverse sources for text-to-video HOI generation. The data exemplifies the setting our anonymizer must handle in deployment: natural lighting, free-form camera motion, varied clothing, and multi-person, multi-object activities. The captions provided in the dataset are derived via a Mixture-of-Multimodal-Experts pipeline \cite{liu2025hoigen,wang2024qwen2,xu2024pllava}, maximizing descriptiveness and accuracy.

We implement classical baselines by performing pixel-level redactions applied inside the human masks $\mathbf{M}$, ranging from performing a Gaussian blur or pixelating to completely blacking out the human regions. We also follow the recommended implementations of EgoBlur~\cite{raina2023egoblur}, DeepPrivacy2~\cite{hukkelaas2023deepprivacy2}, FAMS~\cite{kung2025face}, and FADM~\cite{zwick2024context}.

\subsection{Evaluation protocol}
\label{sec:protocol}
We follow past work~\cite{dinh2026unsafe2safe,zhang2025more} and propose evaluating on privacy, quality, and utility dimensions. Our framework is updated for video evaluation. Each metric takes the triplet $(x,\hat x,y)$---original clip, anonymized clip, caption---and returns a clip-level score.

\paragraph{Privacy.} We propose two metrics that measure how different the anonymized video is from the original:
%\begin{itemize}[leftmargin=*,itemsep=1pt,topsep=2pt]
%\item 
(1) \textbf{Identity Similarity} \cite{dinh2026unsafe2safe}: the cosine similarity between detected faces and their anonymized counterparts (using ArcFace embeddings).
%\item 
(2) \textbf{Text Similarity}: the mean best Levenshtein-ratio match between detected text in the original and the anonymized video.
%\end{itemize}

However, these two metrics do not measure how well a method anonymizes the body. So, we also test the privacy of our model under a body-aware attacker by running CSCI-V~\cite{pathak2025colors} (a state-of-the-art video re-identification model) against the CCVID clothes-changing person re-ID benchmark~\cite{gu2022clothes}. We anonymize a fixed $500$-tracklet subset of the CCVID query split and leave the full gallery un-anonymized, and re-query: a match means the attacker recovered the identity despite anonymization. 

\paragraph{Quality.} For quality evaluation, we employ the widely used video generation evaluation benchmarks VBench~\cite{huang2024vbench} and VBench-2.0~\cite{zheng2025vbench}. From VBench, we report \textbf{Subject Consistency} and \textbf{Background Consistency} (within-clip coherence of the foreground subject and surrounding scene), \textbf{Motion Smoothness} and \textbf{Temporal Flickering} (frame-to-frame stability of motion and pixel intensity), and \textbf{Overall Consistency} (caption alignment). From VBench-2.0, we report \textbf{Human Anatomy} and \textbf{Human Identity}, which respectively measure anatomical plausibility and how stably the generated person persists across frames. We additionally prompt a VLM~\cite{deitke2025molmo} on how likely the video is to be a real-world video and report a \textbf{Realism} score. 

% Requires: \usepackage{makecell} in the preamble
\definecolor{TabBest}{RGB}{254,224,182}    % Okabe-Ito orange (light)
\definecolor{TabSecond}{RGB}{209,229,240}  % Okabe-Ito sky blue (light)
\newcommand{\best}[1]{\cellcolor{TabBest}#1}
\newcommand{\snd}[1]{\cellcolor{TabSecond}#1}

\begin{table*}[t]
\centering
\setlength{\tabcolsep}{3pt}
\renewcommand{\arraystretch}{1.15}
\resizebox{\textwidth}{!}{%
\begin{tabular}{l cc ccc ccc}
\toprule
& \multicolumn{2}{c}{\textbf{Appearance leakage}} & \multicolumn{3}{c}{\textbf{Body re-identification: Standard (SC)}} & \multicolumn{3}{c}{\textbf{Body re-identification: Clothes-Changing (CC)}} \\
\cmidrule(lr){2-3} \cmidrule(lr){4-6} \cmidrule(lr){7-9}
Method & Identity Similarity $\downarrow$ & Text Similarity $\downarrow$ & R-1 $\downarrow$ & R-5 $\downarrow$ & mAP $\downarrow$ & R-1 $\downarrow$ & R-5 $\downarrow$ & mAP $\downarrow$ \\
\midrule\midrule
\multicolumn{9}{l}{\emph{Classical}} \\
EgoBlur (face only)~\cite{raina2023egoblur} & \best{0.0014} & 0.9263 & 100.0 & 100.0 & 99.9 & 86.6 & 91.2 & 87.2 \\
Gaussian blur & 0.0148 & 0.7694 & 50.0 & 60.8 & 53.3 & 40.0 & 51.6 & 43.3 \\
Pixelation & \snd{0.0039} & 0.7549 & 100.0 & 100.0 & 99.8 & 84.8 & 90.6 & 85.9 \\
Blackout & 0.0039 & 0.7349 & 7.6 & 18.4 & 14.2 & \snd{17.4} & \snd{29.6} & \snd{20.7} \\
\midrule
\multicolumn{9}{l}{\emph{Face-only, generative}} \\
FAMS~\cite{kung2025face} & 0.0884 & 0.8396 & 100.0 & 100.0 & 100.0 & 90.8 & 92.0 & 90.1 \\
DeepPrivacy2 (face)~\cite{hukkelaas2023deepprivacy2} & 0.0603 & 0.9166 & 79.1 & 87.3 & 79.0 & 69.4 & 79.2 & 70.4 \\
\midrule
\multicolumn{9}{l}{\emph{Body-only, generative}} \\
DeepPrivacy2 (body)~\cite{hukkelaas2023deepprivacy2} & 0.0309 & 0.7404 & \snd{7.0} & \snd{13.3} & \snd{11.3} & 29.0 & 40.0 & 33.7 \\
FADM~\cite{zwick2024context} & 0.0728 & 0.7961 & 96.2 & 98.7 & 96.7 & 86.4 & 90.4 & 85.9 \\
\midrule
\multicolumn{9}{l}{\emph{Ours (Wan2.1-VACE conditioning)}} \\
\textbf{StructHuman (ours)} & 0.0092 & \snd{0.3593} & \best{3.2}  & \best{3.2} & \best{4.6} & \best{1.4}  & \best{2.0} & \best{2.5} \\
\textbf{StructAll (ours)} & 0.0063 & \best{0.0753} & --- & --- & ---& --- &  --- & --- \\
\bottomrule\bottomrule
\end{tabular}%
}
\vspace{5pt}
\caption{\textbf{Privacy evaluation.} Privacy metrics on HOI-Gen1M~\cite{liu2025hoigen} + body-reidentification performance. Lower is better for all columns ($\downarrow$). Per-column \colorbox{TabBest}{best} and \colorbox{TabSecond}{second} highlighted. For fairness of comparison on pedestrian re-identification, we only report \textbf{StructHuman}.}
\label{tab:privacy}

\vspace{1em}

\centering
\setlength{\tabcolsep}{3pt}
\renewcommand{\arraystretch}{1.15}
\resizebox{\textwidth}{!}{%
\begin{tabular}{l ccccc ccc}
\toprule
& \multicolumn{8}{c}{\textbf{Quality}} \\
\cmidrule(lr){2-9}
Method & \makecell{Subject\\Consistency} $\uparrow$ & \makecell{Background\\Consistency} $\uparrow$ & \makecell{Motion\\Smoothness} $\uparrow$ & \makecell{Temporal\\Flickering} $\uparrow$ & \makecell{Overall\\Consistency} $\uparrow$ & \makecell{Human\\Anatomy} $\uparrow$ & \makecell{Human\\Identity} $\uparrow$ & Realism $\uparrow$ \\
\midrule\midrule
\multicolumn{9}{l}{\emph{Classical}} \\
EgoBlur (face only)~\cite{raina2023egoblur} & 92.18\% & 92.54\% & 96.92\% & 95.10\% & 12.43\% & 83.70\% & 13.20\% & 84.00\% \\
Gaussian blur & 92.02\% & \snd{92.96\%} & \snd{97.41\%} & \snd{95.80\%} & 13.62\% & 75.30\% & 14.49\% & 75.04\% \\
Pixelation & \best{93.01\%} & \best{93.32\%} & 96.44\% & 94.98\% & 13.14\% & 70.15\% & 6.93\% & 60.92\% \\
Blackout & 92.07\% & 92.42\% & \best{97.43\%} & \best{95.96\%} & 11.79\% & 81.13\% & 7.69\% & 69.76\% \\
\midrule
\multicolumn{9}{l}{\emph{Face-only, generative}} \\
FAMS~\cite{kung2025face} & 90.71\% & 90.18\% & 96.61\% & 94.75\% & 12.16\% & 74.31\% & 8.23\% & 87.86\% \\
DeepPrivacy2 (face)~\cite{hukkelaas2023deepprivacy2} & 91.67\% & 91.26\% & 96.77\% & 94.88\% & 11.99\% & \snd{87.34\%} & 11.86\% & 88.45\% \\
\midrule
\multicolumn{9}{l}{\emph{Body-only, generative}} \\
DeepPrivacy2 (body)~\cite{hukkelaas2023deepprivacy2} & 89.16\% & 88.76\% & 95.78\% & 94.03\% & 12.06\% & 76.17\% & 10.72\% & 83.74\% \\
FADM~\cite{zwick2024context} & 88.19\% & 88.48\% & 95.86\% & 94.11\% & 12.02\% & 83.55\% & 10.88\% & 88.00\% \\
\midrule
\multicolumn{9}{l}{\emph{Ours (Wan2.1-VACE conditioning)}} \\
\textbf{StructHuman (ours)} & 92.23\% & 91.52\% & 97.12\% & 95.05\% & \best{16.36\%} & \best{89.74\%} & \best{19.30\%} & \snd{89.49\%} \\
\textbf{StructAll (ours)} & \snd{92.88\%} & 92.35\% & 96.98\% & 94.91\% & \snd{16.11\%} & 86.85\% & \snd{17.35\%} & \best{89.65\%} \\
\bottomrule\bottomrule
\end{tabular}%
}
\vspace{5pt}
\caption{\textbf{Quality evaluation}.  Quality~\cite{huang2024vbench,zheng2025vbench} metrics on HOIGen-1M~\cite{liu2025hoigen} Higher is better for all columns ($\uparrow$). Per-column \colorbox{TabBest}{best} and \colorbox{TabSecond}{second} highlighted.}
\label{tab:quality}
\end{table*}

\paragraph{Utility.} To test whether the anonymized clip still supports downstream tasks, we evaluate Video Question Answering in Section~\ref{sec:hoigen_vqa}.

\subsection{Results on HOIGen-1M}
\label{sec:main_results}
We anonymize a randomly sampled $1{,}000$-clip subset of HOIGen-1M~\cite{liu2025hoigen} and report a qualitative comparison in Figure~\ref{fig:qual}, as well as a privacy and quality comparisons in Table~\ref{tab:privacy} and Table~\ref{tab:quality}.

Face-only baselines leave the body untouched, resulting in high re-identification scores.Full body redactions that destroy the signal entirely (Blackout, pixelation, Gaussian blur) are much more private, but suffer in quality --- they have low Realism scores and low Overall Consistency. Generative inpainting of the body suffers in quality and is still not completely private; FADM is even less private than than the face-only DeepPrivacy2. Interestingly, classical baselines score high in some quality measures. We attribute this to the same redaction operation being performed across frames, trivially achieving high "consistency".

Our \textbf{StructAll} and \textbf{StructHuman} are among the most private, driving re-identification to near chance. The fact that our method outperforms even blackout on re-identification shows that re-generating the entire body with a realistic appearance not only removes identity signal but even misleads re-identification attempts. We lead on Overall Consistency, Human Anatomy, Human Identity, and Realism, in addition to performing the best amongst the generative full body methods in the other quality metrics. Our method is also the only one to consistently anonymize text, a useful capability for many high-sensitivity scenarios.

Classical anonymizers (blackout, pixelation) may be private but collapse on quality. At the same time, anonymizers that may achieve a higher quality (EgoBlur, Gaussian blur) leave the body re-identifiable. Our methods sit near the top on both axes simultaneously, generating videos that are simultaneously private yet high quality.

% Requires: \usepackage{xcolor,colortbl,booktabs,graphicx} and the \best/\snd
% commands defined alongside tab:privacy/tab:quality (TabBest/TabSecond colors).
% \best{...} = best-in-column highlight, \snd{...} = second-in-column.

% Requires: \usepackage{xcolor,colortbl,booktabs,graphicx} and the \best/\snd
% commands defined alongside tab:privacy/tab:quality (TabBest/TabSecond colors).
% \best{...} = best-in-column highlight, \snd{...} = second-in-column.

\begin{table*}[t]
\centering
\setlength{\tabcolsep}{6pt}
\renewcommand{\arraystretch}{1.15}
\resizebox{\textwidth}{!}{%
\begin{tabular}{l cc cc cc}
\toprule
& \multicolumn{2}{c}{\textbf{HOIGen (zero-shot)}} & \multicolumn{2}{c}{\textbf{NExT-QA (zero-shot)}} & \multicolumn{2}{c}{\textbf{MedVideoCap (zero-shot)}} \\
\cmidrule(lr){2-3} \cmidrule(lr){4-5} \cmidrule(lr){6-7}
Method & Acc.\ ($\%$) $\uparrow$ & $\Delta$ & Acc.\ ($\%$) $\uparrow$ & $\Delta$ & Acc.\ ($\%$) $\uparrow$ & $\Delta$ \\
\midrule\midrule
\multicolumn{7}{l}{\emph{Original}} \\
Original (upper bound) & 86.40 & --- & 78.31 & --- & 90.47 & --- \\ % TODO: MedVideoCap upper bounds
\midrule
\multicolumn{7}{l}{\emph{Classical}} \\
EgoBlur~\cite{raina2023egoblur} & \best{86.90} & \best{+0.50} & 77.56 & -0.75 & \best{90.60} & \best{+0.13} \\ % TODO
Gaussian blur & 81.33 & -5.07 & 72.93 & -5.38 & 85.27 & -5.20 \\ % TODO
Pixelation & 78.20 & -8.20 & 73.76 & -4.55 & 85.13 & -5.34 \\ % TODO
Blackout & 73.47 & -12.93 & 73.00 & -5.31 & 74.80 & -15.67 \\ % TODO
\midrule
\multicolumn{7}{l}{\emph{Face-only, generative}} \\
FAMS~\cite{kung2025face} & \snd{86.13} & \snd{-0.27} & \best{77.97} & \best{-0.34} & 89.60 & -0.87 \\ % TODO
DeepPrivacy2 (face)~\cite{hukkelaas2023deepprivacy2} & 85.60 & -0.80 & \snd{77.80} & \snd{-0.51} & \snd{89.80} & \snd{-0.67} \\ % TODO
\midrule
\multicolumn{7}{l}{\emph{Body-only, generative}} \\
DeepPrivacy2 (body)~\cite{hukkelaas2023deepprivacy2} & 73.10 & -13.30 & 73.39 & -4.92 & 78.07 & -12.40 \\ % TODO
FADM~\cite{zwick2024context} & 83.33 & -3.07 & 76.49 & -1.82 & 86.30 & -4.17 \\ % TODO
\midrule
\multicolumn{7}{l}{\emph{Ours (Wan2.1-VACE conditioning)}} \\
\textbf{StructHuman (ours)} & 82.37 & -4.03 & 75.37 & -2.94 & 85.20 & -5.27 \\ % TODO
\textbf{StructAll (ours)} & 81.80 & -4.60 & 75.85 & -2.46 & 84.67 & -5.80 \\ % TODO
\bottomrule\bottomrule
\end{tabular}%
}
\vspace{5pt}
\caption{\textbf{Utility evaluation.} Zero-shot Video Question Answering accuracy with a frozen Qwen3-VL-8B~\cite{bai2025qwen3} on anonymized clips. $\Delta$ is relative to the un-anonymized upper bound (top row); the upper bound is excluded from the per-column ranking. Per-column \colorbox{TabBest}{best} and \colorbox{TabSecond}{second} highlighted.}
\label{tab:vqa}
\end{table*}

\subsection{Utility: zero-shot Video-QA and clinical fine-tuning}
\label{sec:hoigen_vqa}

Anonymization quality only matters if the anonymized clip still answers the question the original clip would have. We evaluate Video Question Answering on three benchmarks: HOIGen-1M~\cite{liu2025hoigen}; NExT-QA~\cite{xiao2021next}, a general-purpose multiple-choice video-QA benchmark; and MedVideoCap-55K~\cite{wang2025medgen}, a caption-annotated medical-video corpus spanning surgery, bedside care, and patient-doctor interactions. We use a vision-language model (VLM)~\cite{bai2025qwen3} to generate multiple choice questions from the provided detailed captions for HOIGen-1M and MedVideoCap-55K.  For evaluation, a frozen VLM is prompted with a question in addition to 8 anonymized frames, and answers are scored by exact-match accuracy. Results are in Table~\ref{tab:vqa}.

\paragraph{Question answering results.} Face-only baselines (EgoBlur, FAMS, DeepPrivacy2-face) sit within $0.9$ points of the un-anonymized upper bound on all three benchmarks, with EgoBlur even slightly exceeding it on HOIGen ($+0.50$) and MedVideoCap ($+0.13$). This is unsurprising --- they edit only the face, leaving the body pixels the VQA model relies on intact. Among methods that actually modify the body, only FADM beats us zero-shot, and the privacy table shows why: FADM's body re-ID is very high, suggesting that it is minimally modifying the video (see Figure~\ref{fig:qual}). Our slight degradation in video question answering accuracy, on the other hand, is exchanged for privacy.

\begin{figure}
  \centering
  \includegraphics[width=\linewidth]{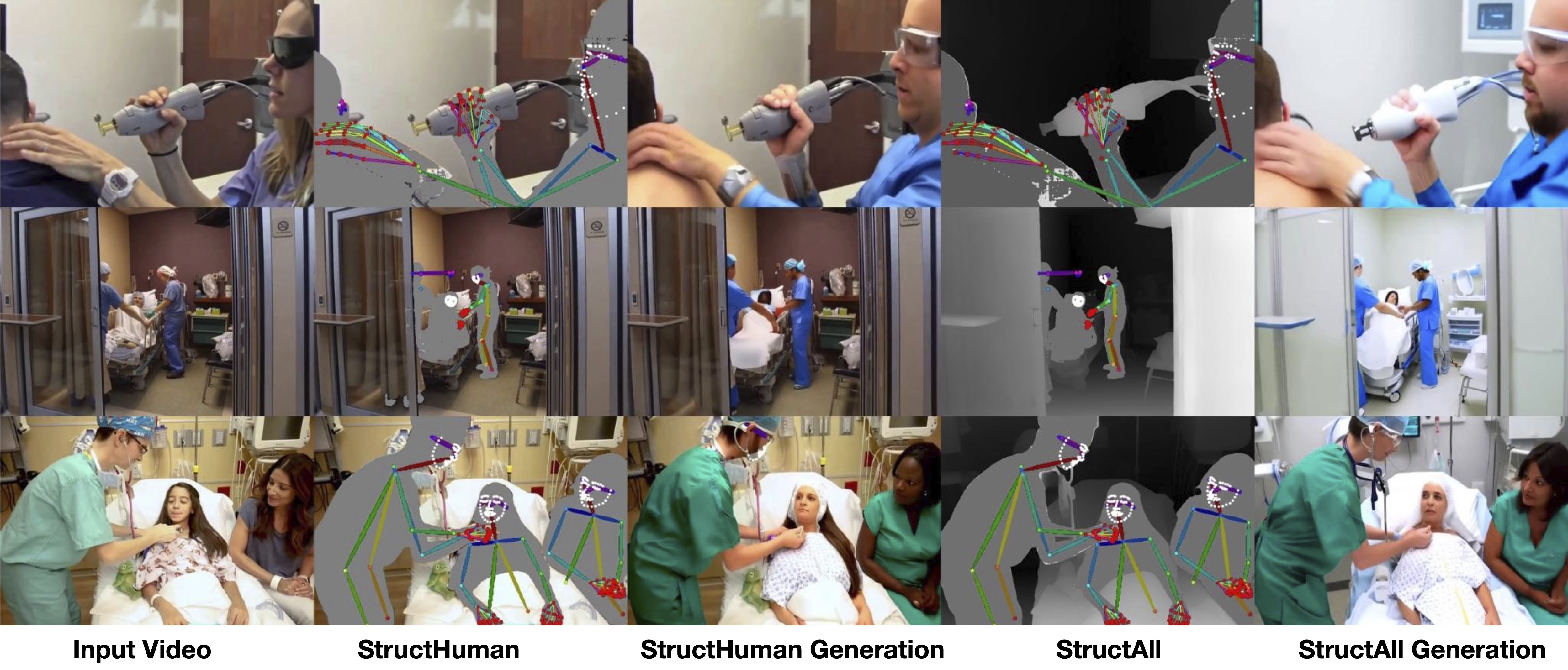}
  \caption{\textbf{Medical video anonymization.} Our model is able to produce high-quality anonymizations of clinical videos from MedVideoCap-55K~\cite{wang2025medgen} whilst retaining the original contexts.}
  \label{fig:med_qual}
\end{figure}

\paragraph{Performance on clinical data.}
Despite being trained only on the general HOIGen-1M~\cite{liu2025hoigen} corpus, our model generalizes to clinical footage out of the box, as shown in Figure~\ref{fig:med_qual} and Table~\ref{tab:vqa}.

This out-of-domain transfer matters because medical video carries a stricter privacy obligation than the in-the-wild footage we trained on. Our results suggest that the structural conditioning and caption carry enough scene context for the model to handle the medical setting without re-training.

% The strongest form of the utility claim is whether one can \emph{train} a downstream medical model on anonymized clips and match the performance of training on raw data. We fine-tune Qwen3-VL-8B on 1,000 videos from MedVideoCap-55K --- anonymized under each regime --- and evaluate all variants on a set of 500 un-anonymized videos (Table~\ref{tab:vqa}, rightmost group). Anonymizing the training data with our method lands at the top of the leaderboard, while keeping the privacy guarantees of full-body re-generation intact.

\begin{figure}
  \centering
  \includegraphics[width=\linewidth]{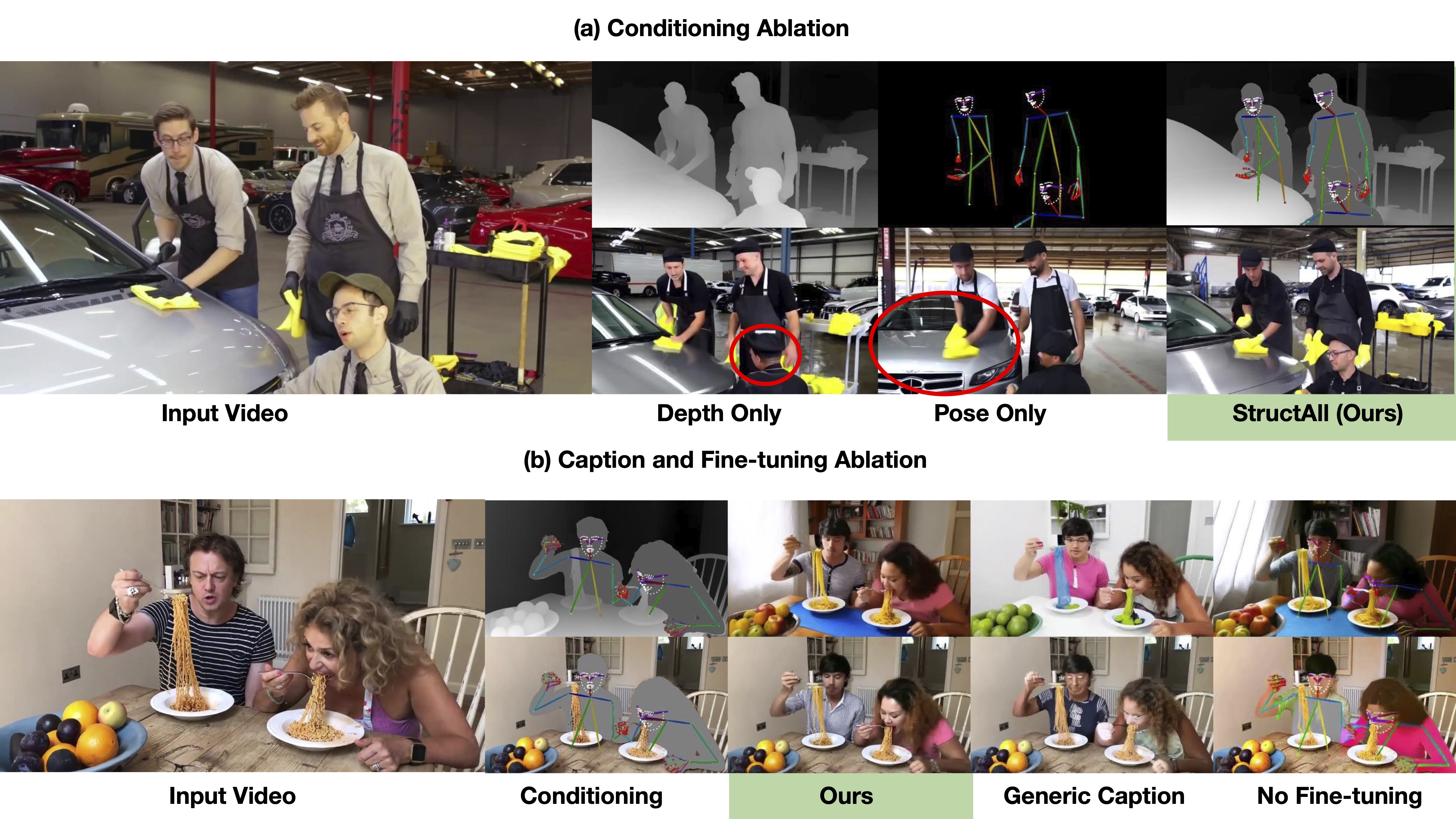}
  \caption{\textbf{Ablation study}. We ablate key aspects of our method. Conditioning on depth-only causes pose ambiguity, while pose-only loses scene geometry (\emph{top}). Without a descriptive caption or fine-tuning, generations degrade in quality and become less consistent with the input video (\emph{bottom}).}
  \label{fig:ablation}
\end{figure}
\subsection{Ablation Study}
\label{sec:ablation}

We ablate the components of our conditioning in Figure~\ref{fig:ablation}. Conditioning on depth alone produces geometrically correct but pose-incorrect generations, while conditioning on pose alone loses scene geometry and results in hallucinations. Using the generic caption ''The video depicts one or more humans performing an activity'' results in lower quality generations that are inconsistent with the original context. The base model without fine-tuning has trouble handling our layered pose and depth conditioning. Combining all of the components of our method is required for optimal performance.
%%%%%%%%%%%%%%%%%%%%%%%%%%%%%%%%%%%%%%%%%%%%%%%%%%%%%%%%%%%%%%%%%%%%%%%%%%%%%%%%
\section{Discussion and Limitations}
\label{sec:discussion}

We present \method, a full-body video anonymization pipeline that regenerates the subject from identity-free structural controls instead of editing original frames, resulting in realistic and temporally consistent anonymizations. We show through our experiments that \method achieves privacy by construction while simultaneously producing realistic high-quality videos that preserve downstream utility. Our work reveals future promise for video anonymization with diffusion-based generative methods and sets a baseline for this relatively new task.

\paragraph{Limitations and Future Work.}
Our work has a few limitations. \emph{(i). Generation time.} A 5-second clip takes $\approx\!2$ minutes to generate on a single 48G NVIDIA L40S GPU---faster than FADM ~\cite{zwick2024context}'s per-frame diffusion, but not real-time. We believe that distillation to a few-step autoregressive student with self-forcing~\cite{huang2025self} is a natural next step, which will also allow for longer video generation. \emph{(ii). Motion and silhouette fidelity}. We currently re-use the source pose verbatim, which preserves gait, a known body-level identifier~\cite{weiss2025privacy}. Perturbing pose sequences and human silhouettes would further extend the privacy guarantee. \emph{(iii). Prompt reliance.} Since Wan2.1 is a text-to-video model by nature, our generations are very reliant on the text prompt, and video context may shift given hallucinations in our captioning model. We study this effect in Section~\ref{app:add} of the appendix.

\paragraph{Broader impacts.} Our work directly enables large-scale release of human-centric video in sensitive areas. Ethical data sharing on a large scale will improve video understanding and generation capabilities of future models, in addition to increasing user trust. We also believe that our architectural guarantee can make our privacy claim auditable --- with a human-in-the-loop approach, privacy professionals can confirm our conditioning inputs (and by extension, the anonymized outputs) are consistent with the input video and contain no private information. Importantly, downstream users should treat generated identities from our method as synthetic, and further work is needed to see if they are provably untraceable to identities from the datasets used to train our model. We continue our discussion of broader societal impacts in Section~\ref{app:impact} of the appendix.

%%%%%%%%%%%%%%%%%%%%%%%%%%%%%%%%%%%%%%%%%%%%%%%%%%%%%%%%%%%%%%%%%%%%%%%%%%%%%%%%

%%%%%%%%%%%%%%%%%%%%%%%%%%%%%%%%%%%%%%%%%%%%%%%%%%%%%%%%%%%%%%%%%%%%%%%%%%%%%%%%
\begin{ack}
This research was supported in part by the National Institutes of Health grant number AG089169. The content is solely the responsibility of the authors and does not necessarily represent the official views of the NIH. 
\end{ack}

{\small
  \bibliographystyle{splncs04}
  \bibliography{main}
}

%%%%%%%%%%%%%%%%%%%%%%%%%%%%%%%%%%%%%%%%%%%%%%%%%%%%%%%%%%%%%%%%%%%%%%%%%%%%%%%%
\clearpage
\appendix
\begin{figure}[th!]
  \centering
  \includegraphics[width=\linewidth]{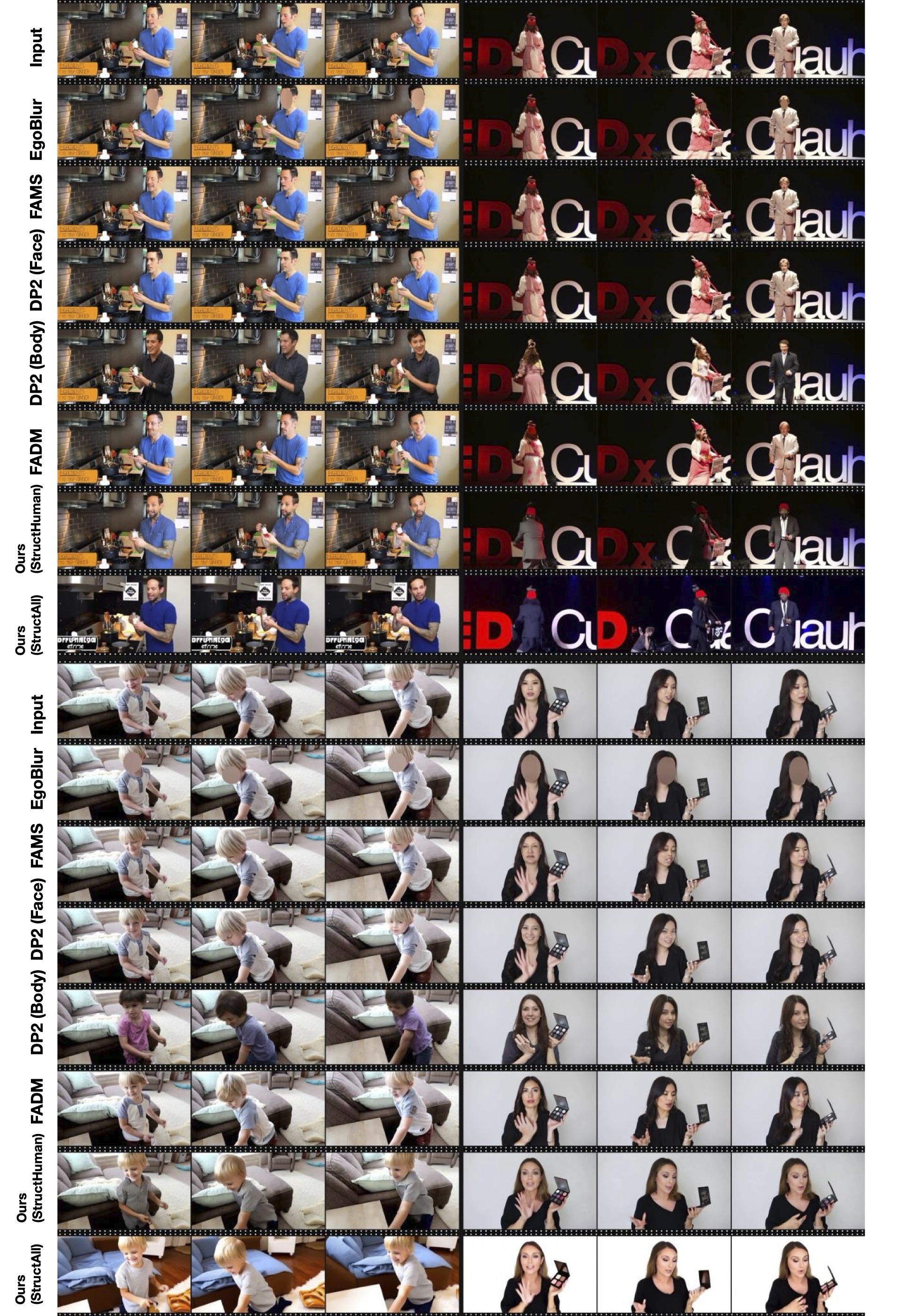}
  \caption{Qualitative comparison on 4 randomly sampled examples from our test subset of HOIGen-1M~\cite{liu2025hoigen}. EgoBlur~\cite{raina2023egoblur}, FAMS~\cite{kung2025face}, and DeepPrivacy2 (Face)~\cite{hukkelaas2023deepprivacy2} do not anonymize the body, resulting in less privacy. DeepPrivacy2 (Body)~\cite{hukkelaas2023deepprivacy2} and FADM~\cite{zwick2024context}'s anonymizations are temporally inconsistent. Our method consistently produces high-quality, temporally consistent anonymizations that can anonymize either just the human or both the human and background.}
  \label{fig:qual}
\end{figure}
\clearpage

\section{Qualitative Results}
\label{app:qual}
We provide additional qualitative results in Figure~\ref{fig:qual} and as a video in the supplementary materials.
\section{Additional Studies}
\label{app:add}
\begin{figure}
  \centering
  \includegraphics[width=\linewidth]{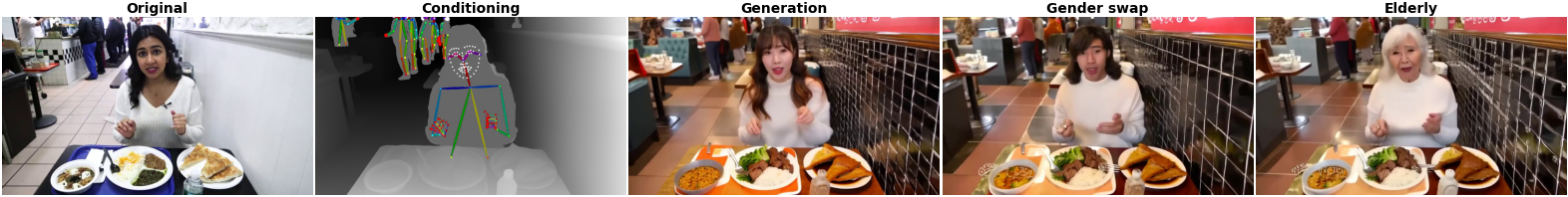}
  \caption{\textbf{Study of prompt reliance.} Our pipeline relies heavily on input captions. Here we show generations resulting from modifying the demographic of the input caption from left to right of "a young woman" (the default) to "a young man" and "an elderly woman". The model generates a similar background and foreground in all cases, but seamlessly swaps out the person.}
  \label{fig:dem}
\end{figure}
We further study our pipeline's reliance on the input caption in Figure~\ref{fig:dem}. By simply editing the demographic in the input caption, we are able to edit the appearance of the generated human. This allows for seamless demographic replacement in addition to anonymization. We note that demographic randomization fundamentally changes the context of the video (i.e. a downstream video question answering task may have differing labels aligned with the input video). Nevertheless, we believe this capability may be promising for certain use cases where demographics are not relevant to the context of the task.

\begin{figure}
  \centering
  \includegraphics[width=\linewidth]{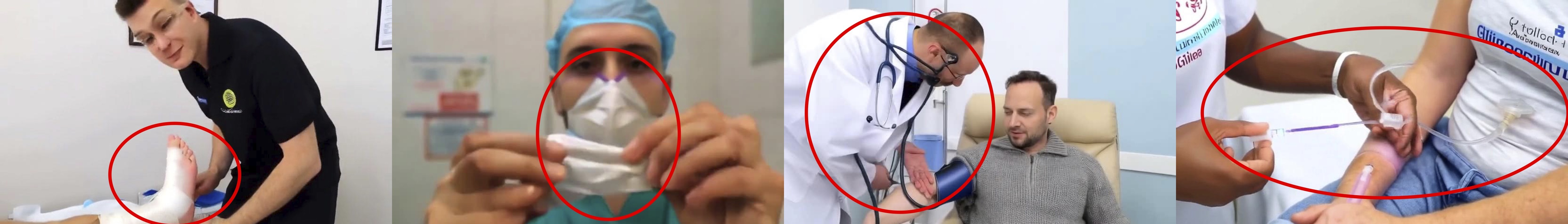}
  \caption{\textbf{Failure cases.} Here we show some failure cases of our model on MedVideoCap-55K~\cite{wang2025medgen}. On the leftmost image, the foot has numerous artifacts and is ambiguously wrapped in a cast. In the next image, a nurse is shown simultaneously putting on a mask and wearing one. The model is not able to generate a doctor wearing a stethoscope, and struggle with generating IVs as well.}
  \label{fig:fail}
\end{figure}

While we believe that our combination of depth, pose, and descriptive text caption provides strong conditioning for many anonymization use cases, our model still inherits many of the failure cases of video diffusion models. We show a few failures we observed on MedVideoCap-55K in Figure~\ref{fig:fail}. The failures largely stem from ambiguous input conditioning and niche contexts. More work is needed to be done to assist video diffusion models in becoming more context-aware.

\section{Dataset details}
\label{app:data}
We train with a subset of the HOI-Gen1M dataset (\url{https://huggingface.co/datasets/HOIGen/HOIGen-1M}), which is licensed under the Apache-2.0 license. We validate with the MedVideoCap-55K dataset (\url{https://huggingface.co/datasets/FreedomIntelligence/MedVideoCap-55K}), NExT-QA dataset (\url{https://github.com/doc-doc/next-qa}), and CCVID dataset (\url{https://opendatalab.com/OpenDataLab/CCVID}). MedVideoCap-55K and CCVID are licensed under the Apache-2.0 license, and NExT-QA is licensed under the MIT license.

We utilize a randomly sampled subset of $7{,}500$ clips from HOI-Gen-1M~\cite{liu2025hoigen} for training. We use $49$-frame clips at $832{\times}480$ and $8$\,fps. We use the same frame sampling regime for our testing on our subsets of CCVID~\cite{gu2022clothes}, NExT-QA~\cite{xiao2021next}, and MedVideoCap-55K~\cite{wang2025medgen}.

\section{Implementation Details}
\label{app:train}

% This section gathers the implementation details for every study reported in the main paper: model training, the CCVID re-identification attack, Video-QA on HOIGen, and the MedVideoCap protocol.

\subsection{Model training and conditioning extraction}
\label{app:train_model}

\paragraph{Training.} For \textbf{StructHuman} and \textbf{StructAll}, we each train a rank-$\rho{=}32$ LoRA on the projection matrices $\{q,k,v,o,\mathrm{ffn}_0,\mathrm{ffn}_2\}$ of every Context Adapter block of a frozen Wan2.1-VACE-1.3B~\cite{wan2025wan,jiang2025vace}. The DiT backbone and the pretrained Context Adapter weights are frozen; only the LoRA factors are trained. We optimize with conditional rectified-flow loss on $(\mathbf{x},\mathbf{c},y)$ triples where $\mathbf{c}\in\{\mathbf{c}_{\mathrm{All}},\mathbf{c}_{\mathrm{Hum}}\}$. We use AdamW at learning rate $10^{-4}$. We train for $14{,}000$ iterations, taking approximately 72 hours with 2 NVIDIA L40S GPUs. At inference, we perform $50$ rectified-flow steps per clip, using a classifier-free guidance scale 5.0. Inference takes approximately $3$ minutes on a single NVIDIA L40S GPU. We utilize the training infrastructure of ModelScope's  DiffSynth-Studio (\url{https://github.com/modelscope/diffsynth-studio}), which is licensed under the Apache-2.0 license.
% TODO(adam): add LoRA $\alpha$ and dropout values.
% TODO(adam): add training batch size (per-GPU and effective).
% TODO(adam): specify training video resolution and clip length used at fine-tuning time (49 frames at 832x480, 8 fps from sec 4.1 — confirm same here, or note if different).
% TODO(adam): specify mixed-precision setting (bf16/fp16/fp32) and whether gradient checkpointing is used.
% TODO(adam): note LR schedule (constant? cosine? linear warmup over how many steps?) and any gradient clipping.
% TODO(adam): describe whether StructHuman and StructAll share one LoRA adapter or use two separate adapters, and how training batches are constructed across the two conditionings (mixed, alternated, or two separate runs).
% TODO(adam): specify the random seed(s) used and whether results are averaged across seeds.
% TODO(adam): note any preprocessing details that matter for reproducibility — gray fill value (g = (128,128,128) — already in method), mask dilation kernel size at training time (if any), pose stroke width, depth normalization scheme.

\begin{figure}
  \centering
  \includegraphics[width=\linewidth]{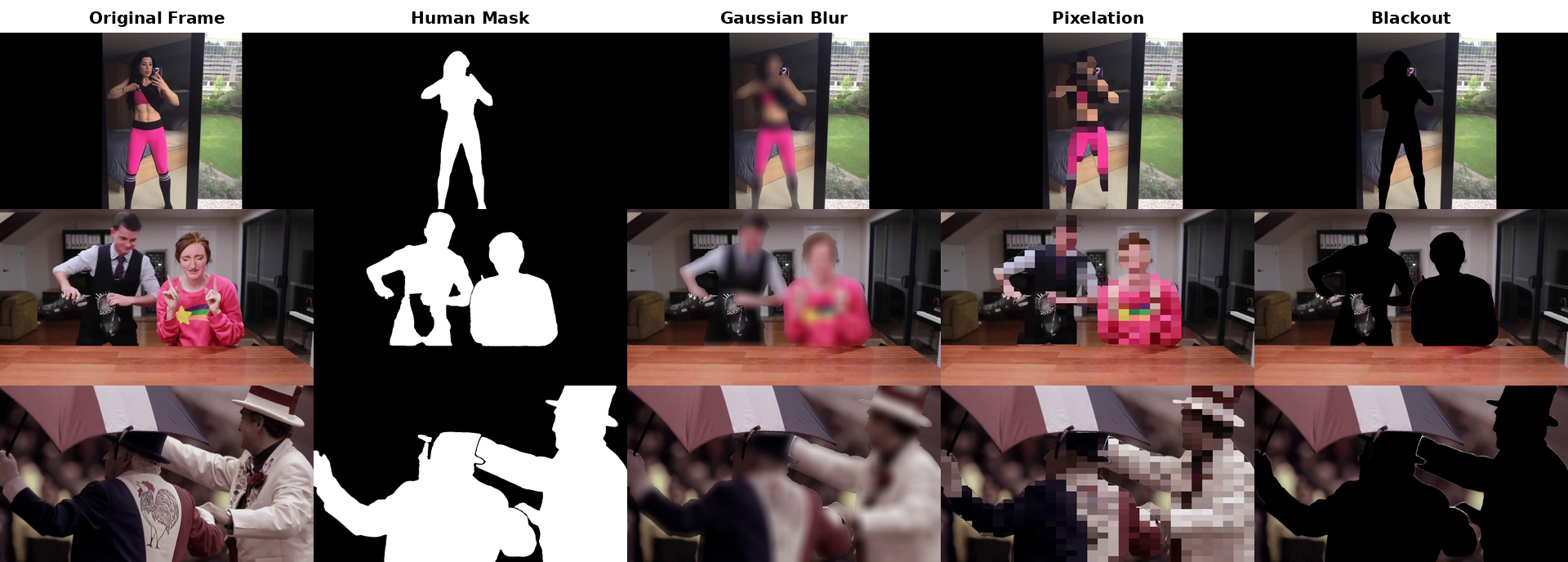}
  \caption{\textbf{Classical anonymization baselines.} Visualization of our implementations of redaction-based anonymization baselines. All approaches use the same SAM3~\cite{carion2025sam} extractions as our methods. }
  \label{fig:classical}
\end{figure}

\paragraph{Conditioning extraction.} We use SAM3~\cite{carion2025sam} for human masks (prompted with "human"), DWPose~\cite{yang2023effective} for the rendered skeleton, and Video Depth Anything~\cite{chen2025video} for monocular depth; and utilize per-clip captions directly from the datasets for HOIGen-1M and MedVideoCap-55K and prompt Qwen3-VL-8B-Instruct~\cite{bai2025qwen3} to generate a detailed caption of the scene when they are not provided. All extractors are frozen.
% TODO(adam): specify the exact model variants / checkpoints used: SAM3 size (e.g. SAM3-L/H), DWPose backbone (e.g. ViTPose-L), Video Depth Anything size (Small/Base/Large).
% TODO(adam): give the exact Qwen3-VL caption prompt used when datasets do not provide a caption (e.g. for CCVID and NExT-QA).
% TODO(adam): specify the resolution / fps at which conditioning is extracted, and any temporal stride.
% TODO(adam): if depth is normalized or rescaled before being used as the StructAll background fill, describe the procedure (per-clip min-max? global range?).
% TODO(adam): if SAM3 masks are post-processed (dilation, hole-filling, temporal smoothing), describe the procedure.

\paragraph{Baselines.}

For our classical redaction baselines ("Gaussian Blur", "Pixelation", "Blackout"), we take the extracted SAM3~\cite{carion2025sam} human masks and conduct the corresponding operations on those pixels. See Figure~\ref{fig:classical} for visualizations.
\begin{itemize}

\item \textbf{Gaussian blur.}
The transformed image is obtained by applying a Gaussian filter to the entire frame:
\begin{equation}
    \mathbf{T} = G_{k, \sigma}(\mathbf{I}),
\end{equation}
where $G_{k,\sigma}$ denotes a Gaussian blur with kernel size $k = 51$ and standard deviation $\sigma = 30.0$.
The large kernel and high sigma ensure that faces and fine-grained details (e.g., text on clothing) become unrecognizable, while rough color and shape information is retained.

\item \textbf{Pixelation.}
The transformed image is produced via a two-step resize operation:
\begin{equation}
    \mathbf{T} = \texttt{resize}_{\uparrow}^{\text{NN}}\!\left(\texttt{resize}_{\downarrow}^{\text{bilinear}}\!\left(\mathbf{I},\; b \times b\right),\; H \times W\right),
\end{equation}
where $b = 30$ is the block size.
The frame is first downsampled to a $30 \times 30$ grid using bilinear interpolation, then upsampled back to the original resolution using nearest-neighbor interpolation.
This produces the characteristic mosaic effect, where each output ``pixel'' covers approximately $\frac{H}{30} \times \frac{W}{30}$ of the original image.
Human regions become coarse color blocks that prevent identification while preserving gross spatial layout.

\item \textbf{Blackout.}
The simplest baseline directly zeroes out all masked pixels:
\begin{equation}
    \mathbf{O} = \mathbf{I} \odot (1 - \mathbf{M}),
\end{equation}
requiring no additional parameters.
This constitutes the most aggressive classical approach: all appearance, texture, and color information within human regions is completely removed, leaving only the silhouette boundary as implicit shape information.
\end{itemize}

For the rest of our baselines, we use the default open-source implementations of EgoBlur~\cite{raina2023egoblur}, DeepPrivacy2~\cite{hukkelaas2023deepprivacy2}, Face Anonymization Made Simple~\cite{kung2025face}, and FADM~\cite{zwick2024context}. 
% TODO(adam): one paragraph documenting how each baseline is run, in enough detail to reproduce.
% Concretely, for each of EgoBlur, DeepPrivacy2 (face), DeepPrivacy2 (body), FAMS, FADM:
%   - which release/checkpoint is used (commit hash or release tag),
%   - the input resolution and fps fed to the baseline,
%   - any non-default hyperparameters (e.g. EgoBlur detection threshold, DeepPrivacy2 truncation, FADM diffusion steps),
%   - whether the baseline operates frame-by-frame or with temporal context,
%   - how the human mask is sourced for body baselines (same SAM3 mask we use for our method? the baseline's own segmenter?),
%   - whether the baseline's outputs are resized/recomposited back into the original frame and how.
% Also document the four classical baselines (Gaussian blur, pixelation, blackout) — kernel size / pixelation block size / fill color, and the fact that they are applied inside the SAM3 mask M.

\subsection{CCVID re-identification attack}
\label{app:reid}

We use CSCI-V~\cite{pathak2025colors} (ICCV 2025, \url{https://github.com/ppriyank/ICCV-CSCI-Person-ReID}) released checkpoint \texttt{ez\_eva02\_vid\_hybrid\_extra\_best.pth} as our attacker. We set the gallery as clean unanonymized tracklets and query with anonymized tracklets, simulating the release of reference footage. We randomly sample 500 query sequences as our test set.
% TODO(adam): the broken \ref{app:metrics} was repaired to \ref{app:metrics_def}, but the symmetric (gallery+query both anonymized) variant is not actually reported anywhere in the current draft — either add a small table for it or remove this sentence.

% \textbf{Protocols.} Standard (\textbf{SC}): all gallery--query pairs allowed. Clothes-changing (\textbf{CC}): same-outfit gallery matches removed, isolating gait/body-shape leakage from appearance. Both reuse the \texttt{eval\_rank} from CSCI's \texttt{test.py} verbatim.

% \textbf{Entry gate.} On \texttt{method=original}, our pipeline reproduces the published CSCI-V numbers (SC R-1 91.7 / mAP 92.2; CC R-1 90.8 / mAP 91.3) within $\pm 0.5$. We do not report any anonymized delta unless this gate passes.
% TODO(adam): re-state the CCVID query-subset size (500 tracklets, mentioned in sec 4.2) and the tracklet-selection procedure (random? stratified by identity? seeded?).
% TODO(adam): describe at what resolution / fps the anonymization is performed before being fed back to CSCI-V (CCVID is pre-cropped at 256 x 128 — does the anonymizer run on the crop directly, or on an upscaled version that is then resized back down?).
% TODO(adam): for each baseline applied to CCVID, note any baseline-specific adaptations (e.g. EgoBlur on a 256 x 128 pedestrian crop with no face detected — is the frame passed through unchanged?).

\subsection{Zero-Shot Video-QA}
\label{app:hoigen_vqa}

\textbf{Question construction.} For each clip we derive three question types from the clip caption: an \emph{action} question (``what activity is the person performing?''), an \emph{object} question (``what objects are the people interacting with?''), and an \emph{interaction} question (``how is the person manipulating the object?''). Multiple choice questions are generated by prompting a frozen  Qwen3-VL-8B-Instruct model with the clip's initial frame and its corresponding detailed caption. Ground-truth answers are extracted from the caption with a small regex+LLM pipeline; clips for which extraction fails are dropped.
% TODO(adam): re-state the HOIGen evaluation subset size (1,000 clips, from sec 4.3).
% TODO(adam): say whether each clip yields one MCQ per type (so 3 MCQs per clip) or one MCQ per clip total, and how the final per-method "Acc.\ (\%)" in tab:vqa is aggregated (mean over questions? over clips?).
% TODO(adam): include or paraphrase the exact MCQ-generation prompt used with Qwen3-VL-8B-Instruct (system prompt + user template).
% TODO(adam): describe how the four distractor options are produced and how they are sanity-checked (reject set? deduplication? exclusivity check vs gold?).

\textbf{Evaluation.} A frozen Qwen3-VL-8B~\cite{bai2025qwen3} is prompted at $448{\times}448$, $8$ uniformly-sampled frames per clip, with a fixed system prompt. We score model answers against the ground truth with exact-match.
% TODO(adam): tab:vqa only reports a single "Acc.\ (\%)" column, but this paragraph claims both BERTScore-F1 and exact-match are reported. Resolve the inconsistency: either drop the BERTScore-F1 mention here, or add a column / supplementary table with BERTScore-F1.
% TODO(adam): include or paraphrase the exact evaluation system prompt / user template used at inference.
% TODO(adam): note the decoding settings (temperature, top-p, max new tokens, greedy vs sampled, seed).

\subsection{MedVideoCap protocol}
\label{app:medvideocap}

\begin{figure}
  \centering
  \includegraphics[width=\linewidth]{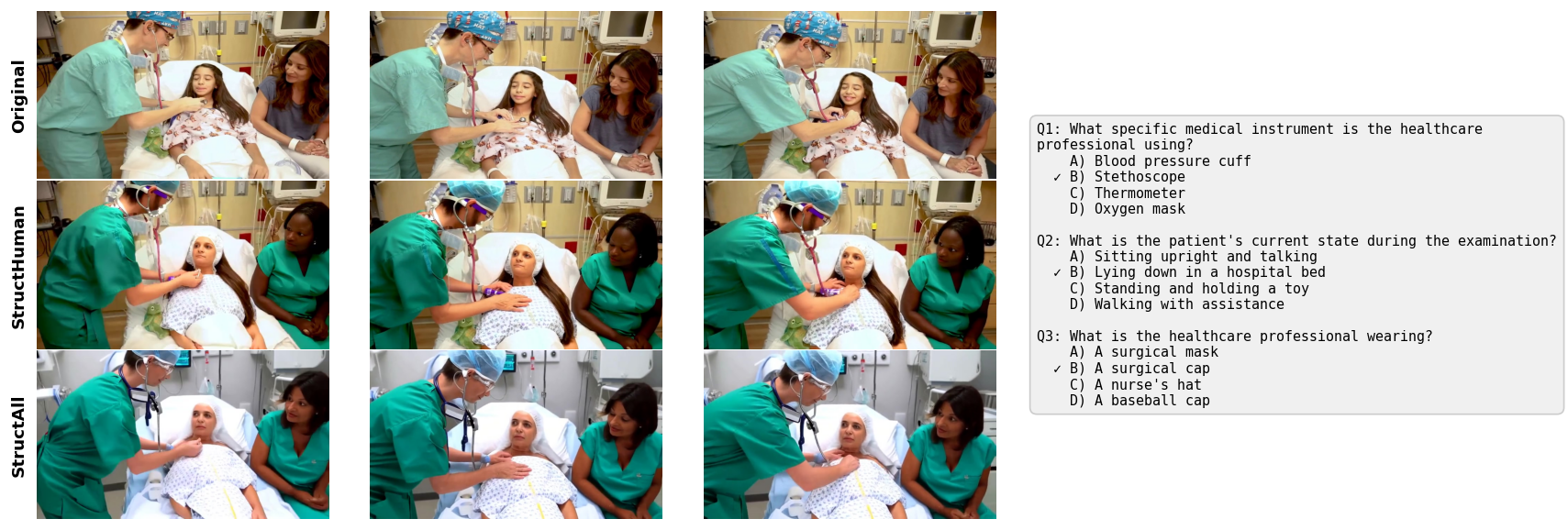}
  \caption{\textbf{Sample generated MedVideoCap-55K questions.} The questions challenge the VLM to reason about the anonymized frames, testing if they remain consistent with the original input video.}
  \label{fig:medical}
\end{figure}

\textbf{Question construction. } We replicate the question construction and evaluation pipeline for the MedVideoCap clips detailed in App.~\ref{app:hoigen_vqa}. We provide sample questions and answers in Figure~\ref{fig:medical}.

\textbf{Subset.} A $1{,}500$-clip randomly sampled MedVideoCap-55K~\cite{wang2025medgen} subset, split into $1{,}000$ clips of train and $500$ clips of test. We filter the dataset to include only "clinical practice videos" whose captions contain mentions of patients and doctors. The resulting dataset only includes full-body clinical context videos, optimal for our video anonymization investigations.

\section{Broader impacts and misuse}
\label{app:impact}
 
\paragraph{Two models, by design.} We propose two variants, \textbf{StructHuman} and \textbf{StructAll}, rather than a single ``one-size-fits-all'' anonymizer. The two condition on different amounts of scene structure --- \textbf{StructHuman} preserves only the human pose and mask, while \textbf{StructAll} additionally preserves a coarse depth backdrop --- and they trade off privacy and utility differently. Forcing a single model would either over-anonymize footage where the surrounding context is itself the signal of interest (e.g.\ a clinical room layout or relevant clinical equipment), or under-anonymize footage where the background carries identifying cues (license plates, signage, identifiable interiors). We allow the deployer match the anonymizer to the downstream task and threat model.
 
\paragraph{Not a deepfake tool, but not risk-free.} Our pipeline is deliberately built without a ground-truth reference frame of any specific person: the conditioning signals (pose, human mask, optional depth) are identity-agnostic, and the model is trained to \emph{break} identity rather than reproduce it. This is a meaningfully different design point from face-swap or talking-head systems that take a target identity as input. That said, the text prompt is still an open channel: a user can describe a specific person --- by name, profession, demographic, or distinctive attribute --- and the underlying video diffusion prior may steer the generated human toward that description. We do not consider this a failure of the anonymization objective (the original identity in the source video is still removed), but it is a possible misuse inherited from the base text-to-video model.
% TODO(adam): if you have a concrete mitigation in mind --- e.g.\ a prompt-side filter, a name-blocklist applied to the user prompt, or a policy recommendation for downstream deployers --- mention it here.
 
\paragraph{Training data and memorization.} The datasets used in this work  are released under permissive licenses compatible with research use.
% TODO(adam): confirm the specific licenses for HOIGen-1M, MedVideoCap-55K, and CCVID --- the original sentence said ``Apache-2.0,'' but Apache-2.0 is a code license, not a data license; please replace with the actual dataset terms (e.g.\ CC-BY, CC-BY-NC, custom research-use license).
However, like any large generative model fine-tuned on video, our LoRA may to some extent memorize training distributions, and a small fraction of generated humans may resemble individuals in the training set. We did not observe verbatim identity reproduction in our qualitative inspection, but we cannot rule it out at scale, and we recommend that downstream users treat anonymized outputs as ``de-identified'' instead of fully new.

 \begin{figure}
  \centering
  \includegraphics[width=\linewidth]{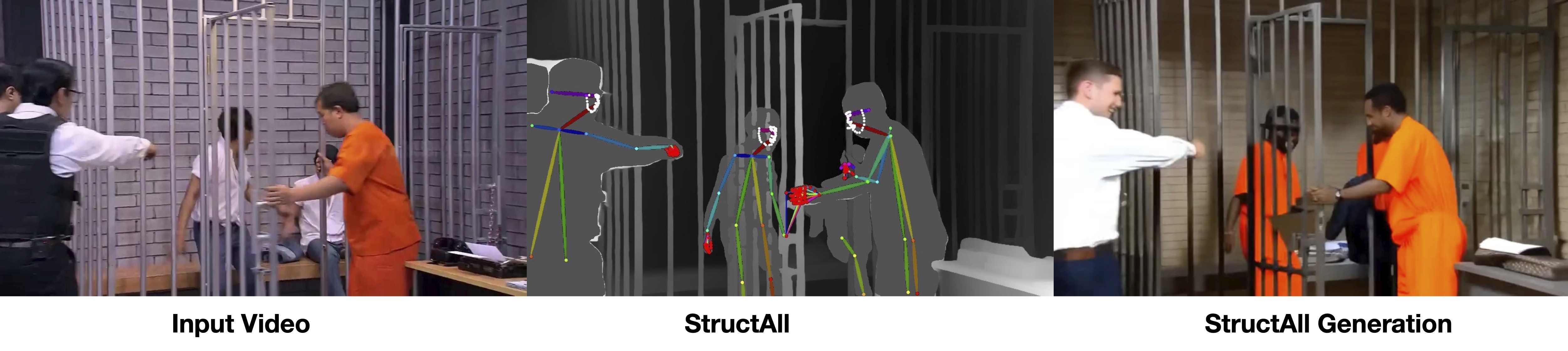}
  \caption{\textbf{Racial bias.} We show one example of bias in our anonymization pipeline. The video generation model assumes Black males are the incarcerated subjects and a White male is the warden.}
  \label{fig:racial}
\end{figure}

\paragraph{Demographic bias.} The model inherits demographic biases from both Wan2.1-VACE's pretraining and our fine-tuning data. One such case is evident when prompted to generate a prison setting --- the model generates Black male subjects as prisoners and white men as wardens, even when the source prompt never explicitly mentions race (Fig.~\ref{fig:racial}).
We attribute this primarily to biases in the base video generation model's training distribution rather than to our approach. One practical mitigation for this task is to inject explicit demographic randomization into the prompt (e.g.\ sample a race/gender token uniformly from a vocabulary before generation --- see Figure~\ref{fig:dem}).
 
\paragraph{Net assessment.} The capabilities our system exposes --- text-and-structure-conditioned video synthesis of humans --- already exist in publicly released video-to-video and text-to-video models. Our contribution is a recipe that bends those capabilities toward privacy preservation rather than identity reproduction, and our threat model (Sec.~\ref{app:reid}) explicitly evaluates the privacy gain rather than asserting it. We therefore believe the marginal misuse risk introduced by this work is small relative to its benefit for downstream privacy-respecting computer vision. We encourage deployers to (a) keep the source footage out of public release, (b) apply prompt-side filtering against named individuals and protected attributes, and (c) prefer \textbf{StructAll} over \textbf{StructHuman} when the surrounding scene is itself sensitive.

% \newpage
% \input{checklist.tex}

\end{document}